\documentclass[10pt,twocolumn,letterpaper]{article}

\usepackage{cvpr}              % To produce the CAMERA-READY version
\usepackage{graphicx}
\usepackage{amsmath}
\usepackage{amssymb}
\usepackage{booktabs}
\usepackage{xcolor}
\usepackage[pagebackref,breaklinks,colorlinks]{hyperref}

\usepackage[capitalize]{cleveref}
\crefname{section}{Sec.}{Secs.}
\Crefname{section}{Section}{Sections}
\Crefname{table}{Table}{Tables}
\crefname{table}{Tab.}{Tabs.}

\newcommand\sw[1]{\textcolor{black}{#1}}

\def\cvprPaperID{*****} % *** Enter the CVPR Paper ID here
\def\confName{CVPRW}
\def\confYear{2023}

\begin{document}
\title{AADiff: Audio-Aligned Video Synthesis with Text-to-Image Diffusion}
\author{Seungwoo Lee$^1$ \and Chaerin Kong$^1$ \and Donghyeon Jeon$^2$ \and Nojun Kwak$^1$ \and
$^1$\text{Seoul National University, }$^2$\text{NAVER}\\
{\tt\small\{seungwoo.lee,vetzylord,nojunk@snu.ac.kr\}, donghyeon.jeon@navercorp.com} 
}
\maketitle
%%%%%%%%% ABSTRACT
\begin{abstract}
Recent advances in diffusion models have showcased promising results in the text-to-video (T2V) synthesis task. However, as these T2V models solely employ text as the guidance, they tend to struggle in modeling detailed temporal dynamics. In this paper, we introduce a novel T2V framework that additionally employs audio signals to control the temporal dynamics, empowering an off-the-shelf T2I diffusion to generate audio-aligned videos. We propose audio-based regional editing and signal smoothing to strike a good balance between the two contradicting desiderata of video synthesis, \textit{i.e.,} temporal flexibility and coherence. We empirically demonstrate the effectiveness of our method through experiments, and further present practical applications for content creation.
\end{abstract}
\begin{figure*}[t]
  \centering
  \includegraphics[width=0.75\linewidth]{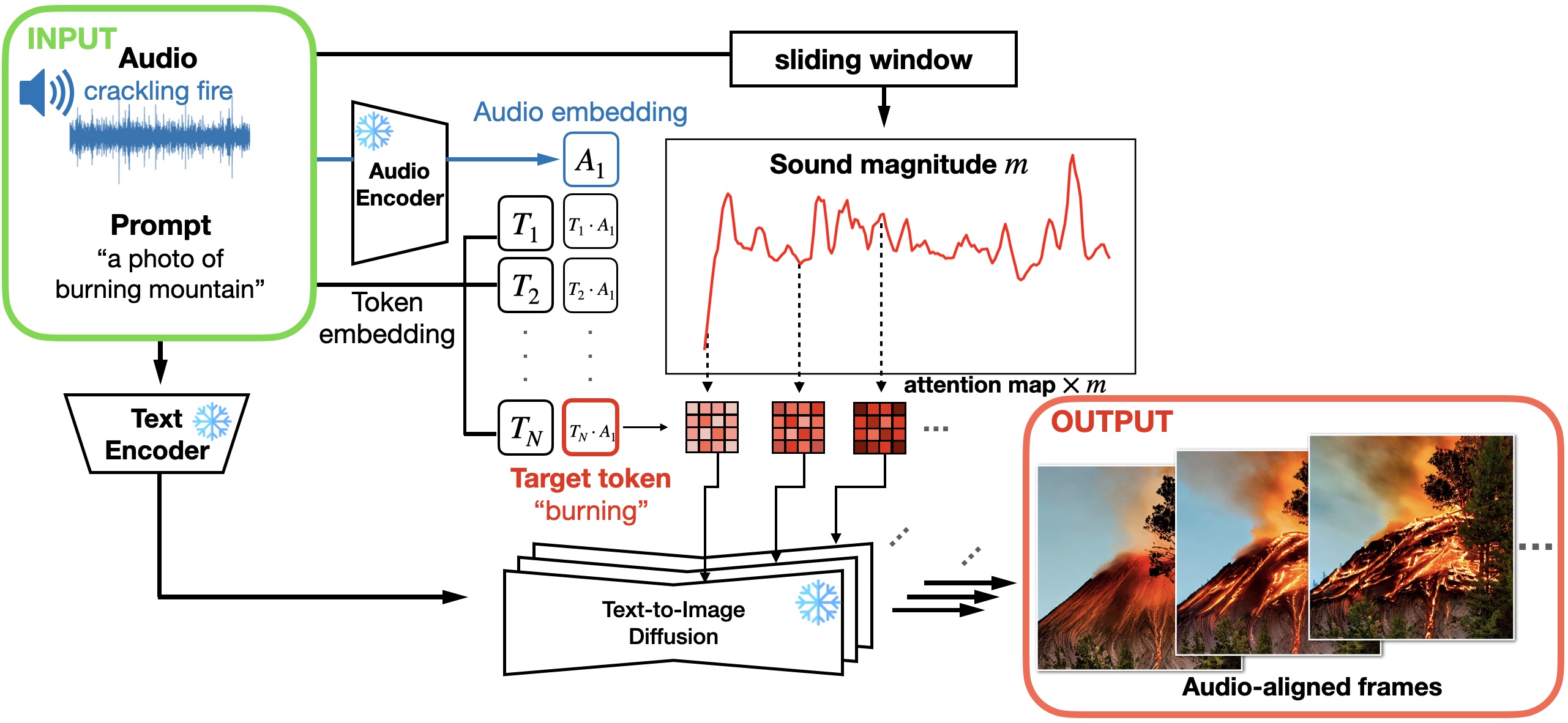}
  \caption{Method overview. \sw{Given an audio signal and text prompt, each is first embedded by the audio encoder and the text encoder, respectively. Text tokens with the highest similarities are chosen and used for editing images with \cite{hertz2022prompt}, where the smoothed audio magnitude controls the attention strength.}}
  \label{fig:method}
\end{figure*}
%%%%%%%%% BODY TEXT
\section{Introduction}
\label{sec:int}

\sw{
Text-to-Image diffusion models~\cite{rombach2022high, ramesh2022hierarchical, saharia2022photorealistic, nichol2021glide, zhang2023adding} have demonstrated unprecedented success across a wide spectrum of generative applications~\cite{hertz2022prompt, gal2022image, ruiz2022dreambooth, poole2022dreamfusion, brooks2022instructpix2pix}, with far\sw{-}reaching influence on not only our research community but the general industry and public. Recently, text-to-video models~\cite{ho2022video, ho2022imagen, wu2022tune, khachatryan2023text2video, molad2023dreamix, qi2023fatezero, ceylan2023pix2video, liu2023video} began to deliver promising results, further expanding our output modality along the temporal axis. However, as these approaches are still in their infantile stage, several limitations render these frameworks yet incomplete. For example, as these models solely rely on the text prompt for guiding the entire generative process, they tend to struggle in modeling detailed temporal dynamics.  Moreover, as their outputs lack accompanying audio, they are more or less closer to \sw{animated} GIFs than proper videos, and even in those cases where we have the audio \textit{a priori}, it is not straightforward to synchronize the video output with this additional condition.
}

\sw{
In order to overcome these limitations, we propose to explicitly incorporate the audio modality, one of the most accessible data sources along with text and image, into our conventional text-to-image pipeline to accomplish a more controllable temporal extension. Specifically, given an off-the-shelf text-to-image diffusion model such as stable-diffusion\footnote{https://github.com/CompVis/stable-diffusion}, we use both the text and the audio to guide video synthesis, where the former focuses on visualizing the scene semantics while the latter is more responsible for the fine-grained control of the temporal dynamics. \sw{At} the high level, we frame video generation as a sequence of image translations starting from a base image (which can be either synthetic or real), and thus employ a well-performing image editing method such as prompt-to-prompt~\cite{hertz2022prompt} to gradually alter the images in compliance with our input conditions.
}

\sw{
Our key contributions can be briefly summarized as:
\begin{itemize}
    \item To our \sw{best} knowledge, we are the first to employ the combination of text and audio to guide \sw{a} diffusion model for video synthesis.
    \item We propose a simple yet effective framework for empowering text-to-image models for audio-synchronized video generation with no need for additional training or paired data. 
    \item We offer promising applications in contents\sw{-}creation by leveraging our capacity to produce videos that are in sync with the audio input.
\end{itemize}
}

\sw{
As further demonstrated in \cref{sec:eva}, our framework lays the foundation for a variety of captivating applications. For instance, a media creator could use public sound sources to produce a short-form video while manipulating the scene composition and appearance with different text prompts. When combined with image inversion techniques like Null Inversion~\cite{mokady2022null}, we can animate still images with audio correspondence, producing more immersive audio-visual contents. Since our framework is orthogonal to its model components, we can continuously benefit from the advances in generative models that \sw{are} ardently taking place. We currently build our method upon stable diffusion, allowing high-quality video synthesis at $512\times512$ scale.
}

\section{Method}
\label{sec:met}

\sw{
In this section, we first introduce preliminaries for our work, namely CLAP~\cite{elizalde2022clap} and LDM~\cite{rombach2022high}, and then present our \textbf{A}udio-\textbf{A}ligned \textbf{Diff}usion framework (dubbed \textbf{AADiff}). The overall pipeline is outlined in \cref{subsec:aadiff}, and we give details on attention-map-based image control and signal smoothing in \cref{subsec:p2p} and \cref{subsec:window}, respectively.
}

\subsection{Preliminary}
\noindent
\sw{ 
\textbf{Contrastive Language-Audio Pretraining (CLAP)}\cite{elizalde2022clap} integrates text and audio by employing two encoders and contrastive learning as done in CLIP~\cite{radford2021learning}. \sw{A version of} CLAP, pre-trained with a huge amount of audio-text pairs, has achieved state-of-the-art performance in multiple zero-shot prediction tasks, \sw{which we employ in our work.}
}

\noindent
\sw{
\textbf{Latent Diffusion Model (LDM)}\cite{rombach2022high} is a compute-efficient diffusion model that uses Variational Auto-Encoder~\cite{kingma2013auto} to first map pixel values to latent codes and applies sequential denoising operations in the latent space. In this work, we employ Stable Diffusion, a family of latent diffusion model that is trained with a vast amount of image-text pairs.
}

\subsection{Audio-aligned Diffusion}
\label{subsec:aadiff}

\sw{Our goal is to generate a video that corresponds to the prompt \sw{upon which fine-detailed dynamic effects are added} \sw{based on} the sound.} \sw{
To this end, we employ three public pretrained foundation models: the text encoder, the audio encoder and the diffusion backbone. For the text encoder and the diffusion generator, we use Stable Diffusion coupled with CLIP~\cite{radford2021learning} as it is a publically available text-to-image model that showcases state-of-the-art performance. CLAP is used to produce the audio embedding, which highlights the \textit{top-k} text tokens upon pairwise similarity. With these text tokens of interest, we obtain the spatial attention map as in \cite{hertz2022prompt}, which will be illustrated in \cref{subsec:p2p}. For relatively simple operations, we generally choose $k=1$.
}

\subsection{Local Editing with Attention Map Control}
\label{subsec:p2p}
\sw{
Prompt-to-prompt\cite{hertz2022prompt} demonstrates a simple yet effective method for text-driven image editing through attention map control.} 
\sw{
As we execute video synthesis with a series of image translations, we apply \cite{hertz2022prompt} with audio-queried \textit{top-k} text tokens for local semantic editing. To model the temporal dynamics with the audio signal, we take the \sw{magnitude of the input audio along the time axis} and use it as a multiplier that controls the \textit{strength} of image editing at each time frame. Specifically, the audio magnitude is multiplied to the attention map between target text tokens and the image, resulting in sharp changes in the highlighted region when the audio signal is strong. This constrains the output video to be in sync with the audio signal, giving our model, AADiff, its namesake.
Note that as we follow \cite{hertz2022prompt} for image editing, this is done at each denoising step. 
}

\subsection{Smoothing Audio with Sliding Window}
\label{subsec:window}
\begin{figure}[t]
  \centering
  \includegraphics[width=0.9\linewidth]{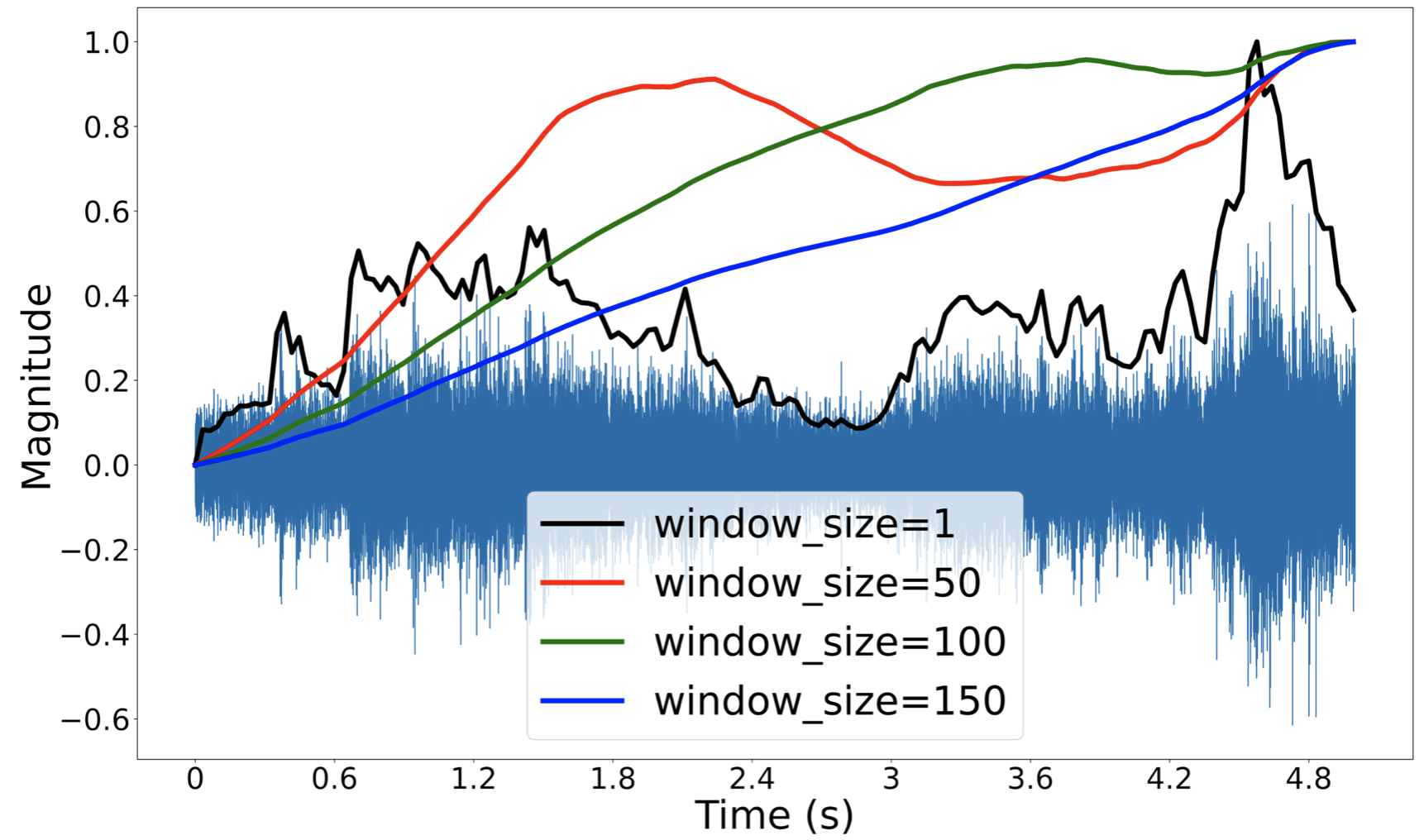}
  \caption{\textbf{Variable sliding window.} A small window size effectively captures dynamic changes, such as thunder, while a larger window size \sw{excels at representing gradual transitions, such as wildfire spreading. This hyperparameter allows the content creator to flexibly control the temporal dynamics of the video.}
}
  \label{fig:sliding}
\end{figure}
\sw{
Using audio magnitude as the guidance signal offers temporal flexibility, but we empirically find that utilizing the raw value for each time frame leads to overly unstable output. Taking the wildfire as an example, the fire would greatly diminish and reignite according to the fluctuations of the audio signal, overall making the output unnatural. In order to overcome this problem, we simply apply a sliding window of size $s$ on the audio magnitude. This smooths the variation in the audio signal and helps producing videos with more natural and coherent dynamics. The effect of varying window size is illustrated in \cref{fig:sliding}.
}

\section{Evaluation}
\label{sec:eva}
\sw{
In this section, we present qualitative and quantitative results to demonstrate the effectiveness of our framework for audio-synchronized video generation. For reproducibility and benchmarking purpose, we run experiments using the audio\sw{s} from the publically available ESC-50\footnote{https://github.com/karolpiczak/ESC-50} test set. Please refer to the supplementary for video samples.
}

\subsection{Audio-aligned Video Synthesis}

\begin{figure}
  \centering
    \includegraphics[width=0.95\linewidth]{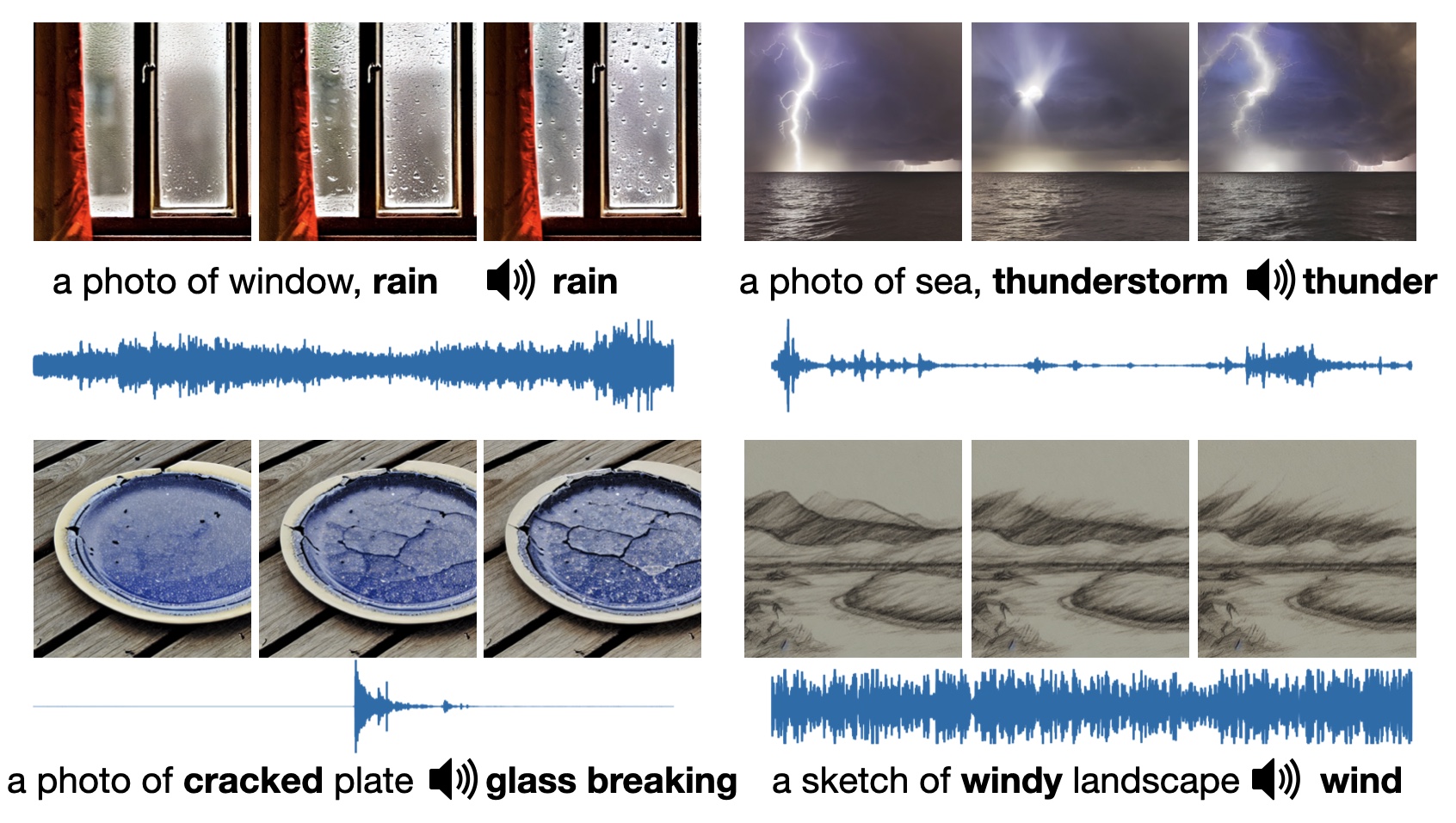}
  \caption{Qualitative result \sw{for different sound sources. Consult the supplementary to check the synchronization between the audio and the output video.}}
  \vspace{-4mm}
  \label{fig:main}
\end{figure}

\sw{
We first test our framework's capacity to output audio-aligned video samples. \cref{fig:main} delivers samples generated with guiding audio of different classes. We can see that our method not only successfully employs the off-the-shelf text-to-image diffusion model for video generation, but further exploits the audio signal to control the temporal dynamics. The sample with \textit{thunderstorm}, for example, is accurately in sync with the audio signal with multiple thunderclaps, getting bright and dark at the right moments (see the video in supplementary).
}

\begin{figure}[t]
  \centering
  \includegraphics[width=0.95\linewidth]{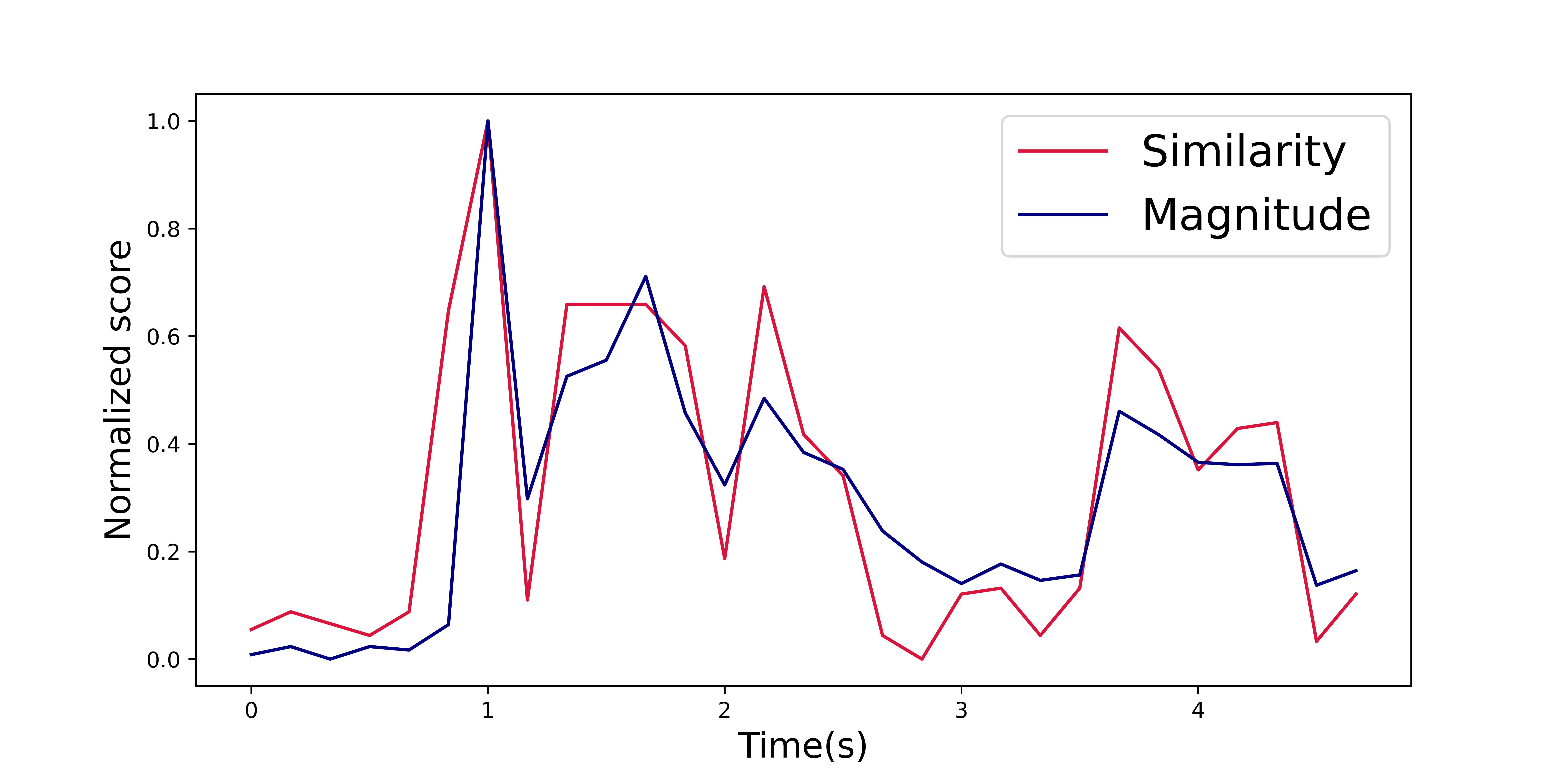}
  \caption{CLIP similarity and \sw{audio} magnitude. \sw{These two values move in unison, indicating that our model faithfully reflects the audio dynamic in the video semantic.}}
  \label{fig:clip}
\end{figure}

\sw{
We further validate this with quantitative measures. We note that quantitative evaluation for text-guided generative models is typically challenging due to the complexity of the task, and no metric introduced so far~\cite{saharia2022photorealistic, rombach2022high} is without flaws. In order to capture our framework's ability to generate \textit{audio-synchronized video}, we present CLIP similarity scores between the textual prompt and synthetic image frame at multiple time steps. The intuition is, when the audio signal is strong, the semantic\sw{s} represented by the audio must also be emphasized, resulting in \sw{a} high CLIP similarity score and vice versa. \cref{fig:clip} confirms that this relation holds in our framework\sw{;} both \sw{the} CLIP score and the audio signal strength \sw{change} in harmony.
}

\begin{figure}[t]
  \centering
  \includegraphics[width=0.88\linewidth]{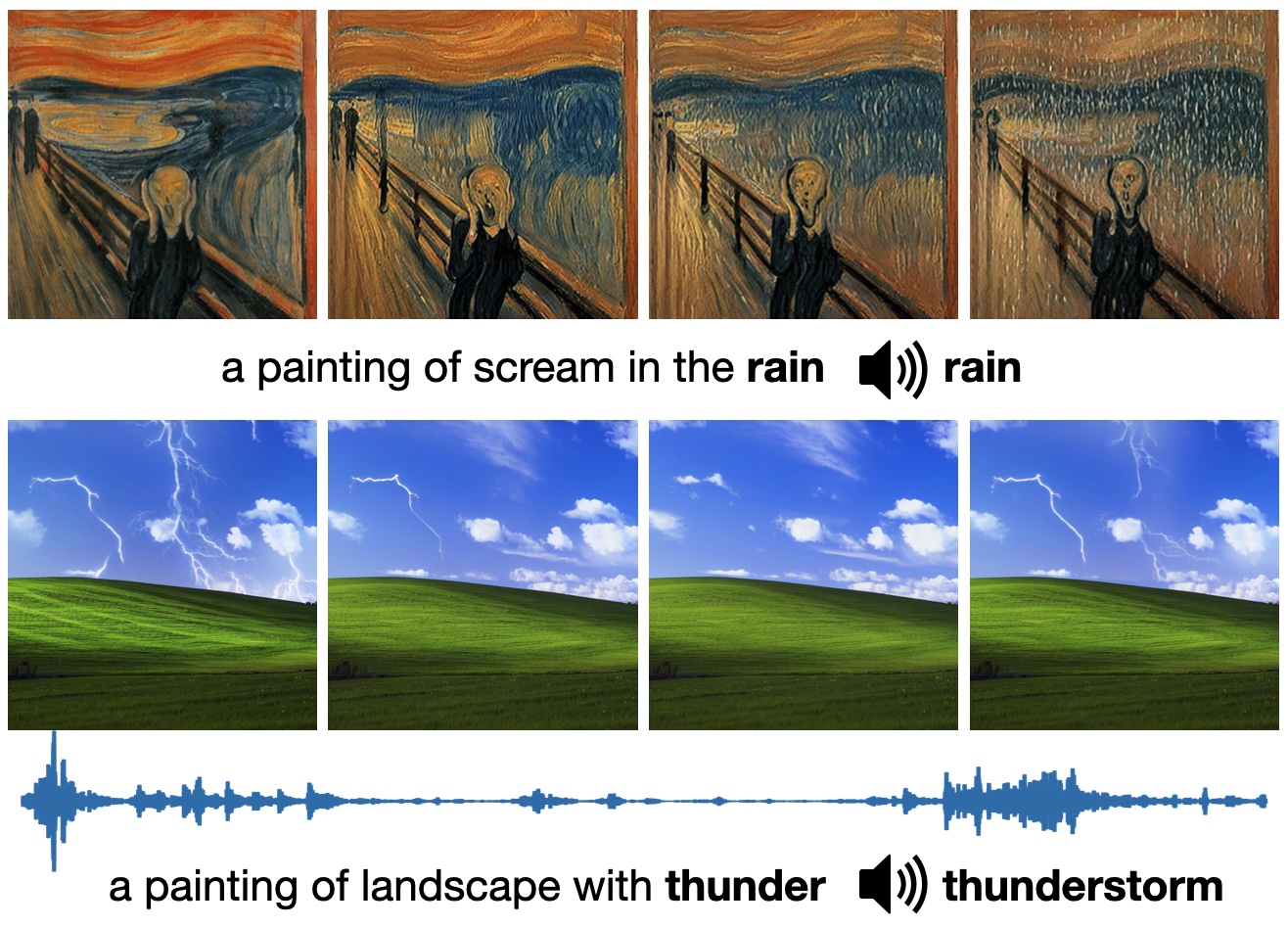}
  \caption{Qualitative results with Null-inversion. \sw{Our method can be used to combine real images and audio sources to create more immersive audio-visual contents.}}
  \vspace{-4mm}
  \label{fig:null}
\end{figure}

\subsection{Animating Still Images}

\sw{
We explore another possible application of animating real images. As numerous works~\cite{mokady2022null, wallace2022edict, song2020denoising} have suggested ways to invert an existing image to the latent space and apply semantic editing operations, we can readily take advantage of them to generate video outputs aligned with any audio signal. Looking at \cref{fig:null} row 1, for example, we can visualize what it would be like if it suddenly started to rain during the man's scream by simply feeding \sw{the} raining sound. Likewise, we see \textit{the bolt out of the blue} in the iconic computer art when the thunderclap noise is added. Note that we consistently obtain videos in sync with the input audio condition.
}

\subsection{Effect of Sliding Window}

\begin{figure}[t]
  \centering
  \includegraphics[width=0.78\linewidth]{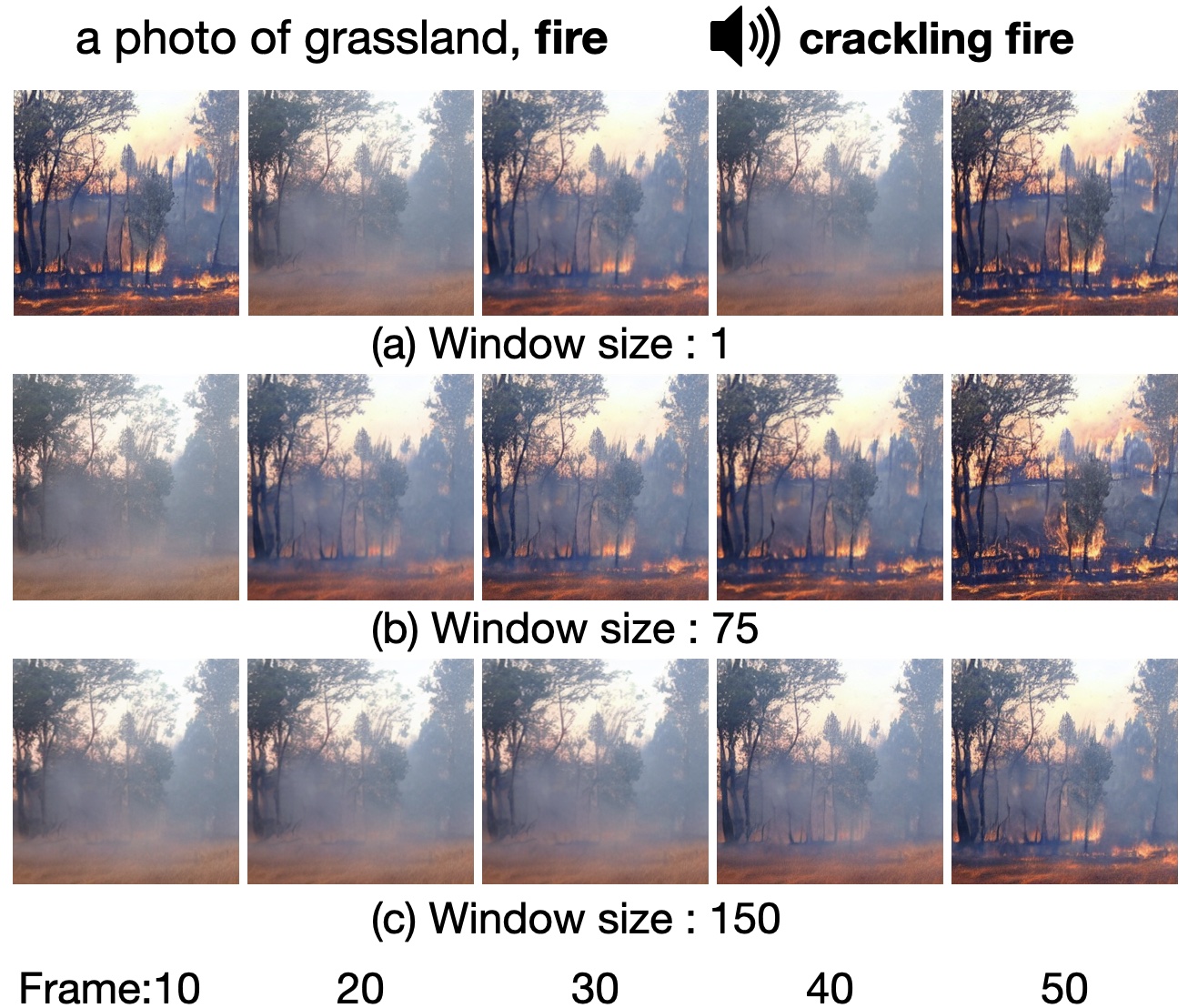}
  \caption{Window size analysis. \sw{No window ($s=1$) leads to excess fluctuation, while infinite window ($s=150$) overly limits the temporal dynamics. We find sweet spots in the mid-range.}}
  \label{fig:window}
\end{figure}

\sw{
In this \sw{part}, we present a brief qualitative ablation on the use of sliding windows. Referring to \cref{fig:window}, \sw{each of the three rows on the left panel} represents no window, window of size 75, and infinite window (note that a 5-second video with 30 fps yields 150 frames total). With no window, the temporal dynamic is too unstable, resulting in temporally inconsistent outputs. When we apply the infinite window, on the other hand, the video is dragged by excessive momentum, producing samples similar to still images. We find a sweet spot in the middle, where a certain level of temporal consistency is guaranteed without overly hurting the dynamic flexibility.
}

% Window Size Analysis

\begin{figure}[t]
  \centering
  \includegraphics[width=0.9\linewidth]{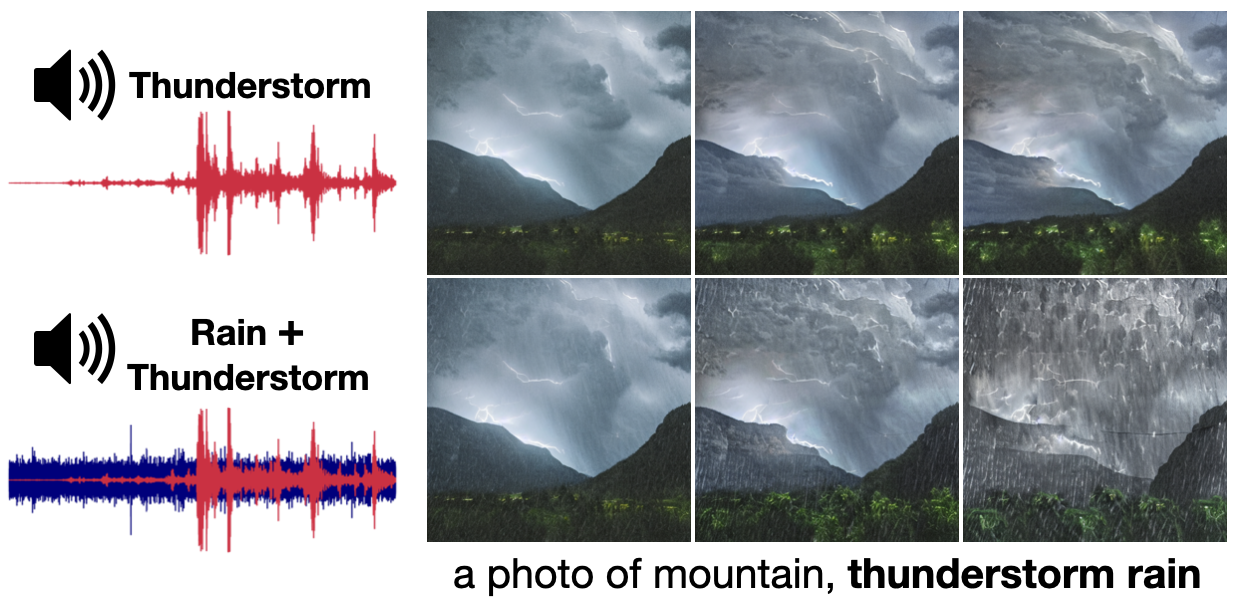}
  \caption{\sw{Video synthesis from \sw{multi} audio \sw{signals}. AADiff \sw{can utilize} different semantics \sw{mixed} in the audio input and produces videos that naturally combine these concepts.}}
  \vspace{-4mm}
   %Multi sounds
  \label{fig:multi}
\end{figure}

\subsection{Further Analyses}

\begin{figure}[t]
  \centering
  \includegraphics[width=0.72\linewidth]{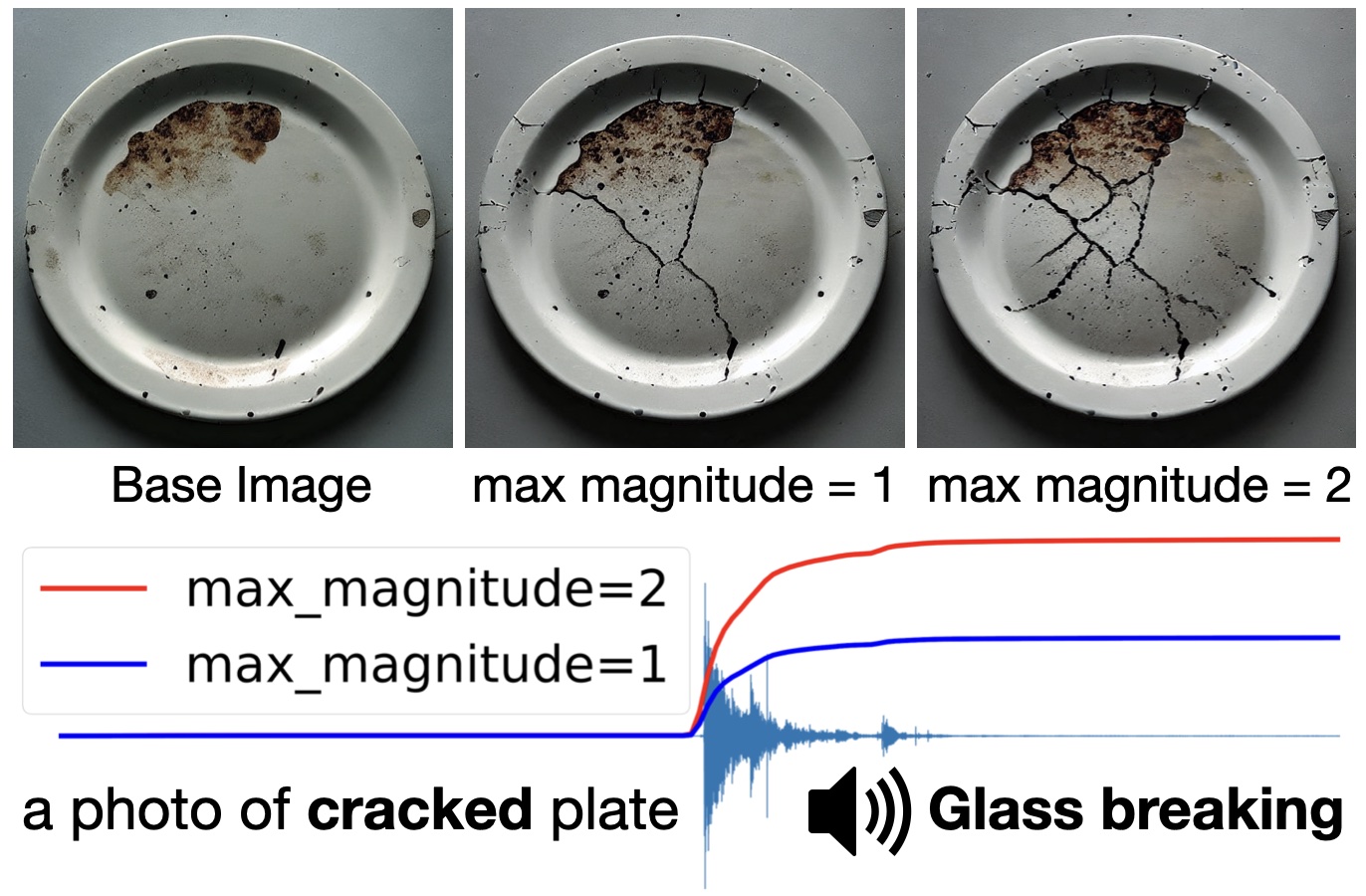}
  \caption{\sw{By differing the audio magnitude, we can further control the degree of deformation.}}
  \label{fig:magnitude}  
\end{figure}

\sw{
Lastly, we study AADiff in different settings and present our findings. We first \sw{use} multiple sound sources and observe our model's generative capacity under this mixed signal. \cref{fig:multi} demonstrates that given a \sw{multi-class} audios, AADiff reflects each component and turn\sw{s} them into a well\sw{-}synchronized video. \cref{fig:magnitude} suggests deeper controllability by changing the magnitude of the audio signal. We clearly observe that with \sw{a} stronger audio input, the output semantic is deformed to a greater extent.
}

\begin{figure}[t]
  \centering
  \includegraphics[width=0.82\linewidth]{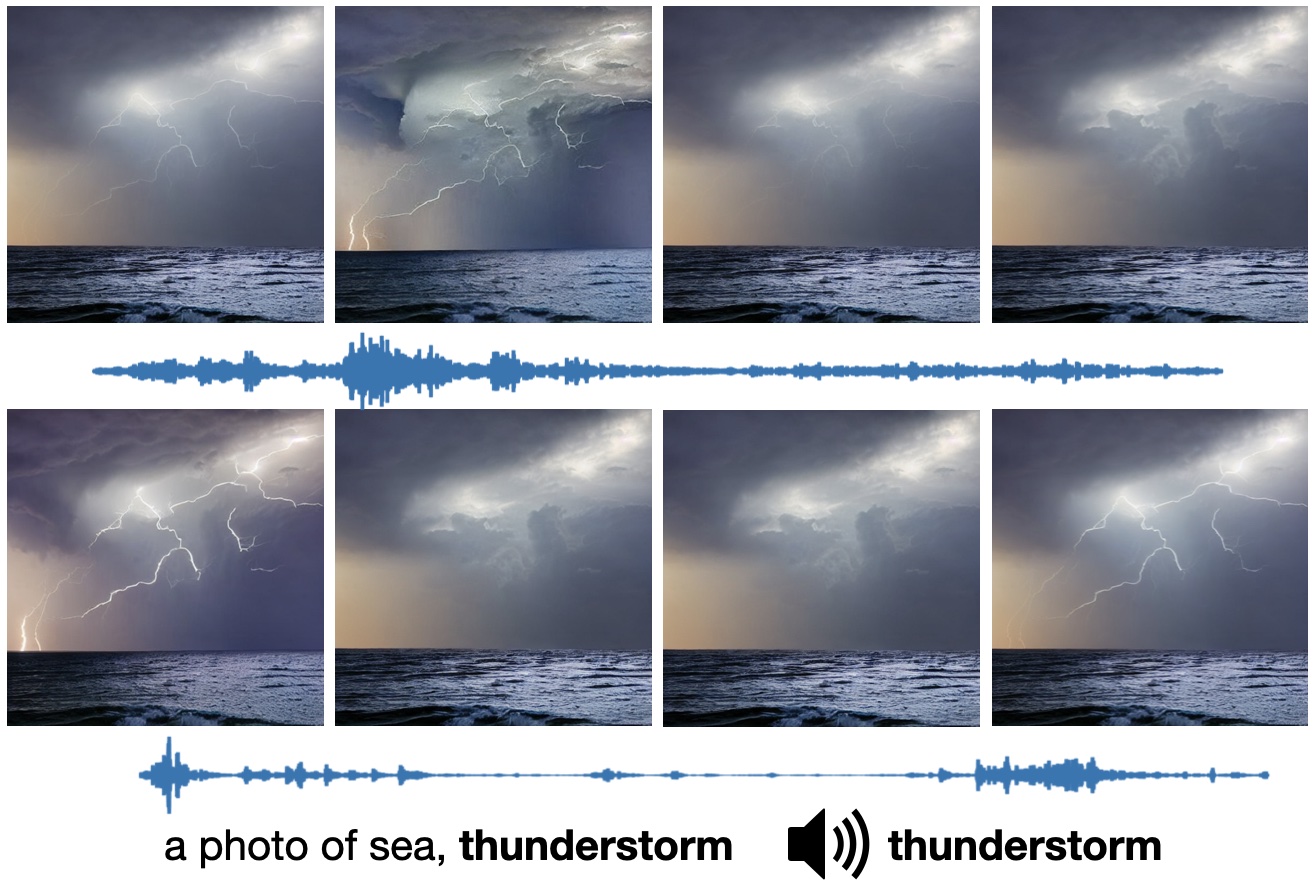}
  \caption{\sw{Unlike conventional text-to-video models, AADiff incorporates the temporal dynamics of the audio input, generating different videos when different sounds of the same class are given.}}
  \label{fig:different}
\end{figure}

\sw{
In \cref{fig:different}, we take a closer look at how our framework incorporates the audio signal. Despite being given the same class of sound (\textit{i.e., thunderstorm}), as the audio contents differ, the output videos also have different visual dynamics. This distinguishes our method from purely text-driven video synthesis methods~\cite{khachatryan2023text2video, wu2022tune, qi2023fatezero}, which typically lack the means to control the temporal dynamics in a fine-grained manner.
}

\section{Conclusion}

\sw{
In this work, we propose a novel framework that takes both text and audio as inputs and generates audio-synchronized videos. As it requires no additional training or paired data of any kind, it can take full advantage of the state-of-the-art multimodal foundation models in a straightforward manner. We hope our work ignites intriguing future works for contents creation.
}
\section{Acknowledgement}
\sw{ This work was funded by Korean Government through NRF grant 2021R1A2C3006659 and IITP grant 2022-0-00320.}
%%%%%%%%% REFERENCES
{\small
\bibliographystyle{ieee_fullname}
\bibliography{egbib}
}

\end{document}